\documentclass[runningheads]{llncs}
\usepackage[T1]{fontenc}

\usepackage{graphicx}
\usepackage[x11names]{xcolor} 
\usepackage{amssymb}
\usepackage{hyperref}
\usepackage{makecell} 
\usepackage{comment}
\definecolor{rolecolor}{RGB}{255,225,225}         
\definecolor{instructioncolor}{RGB}{204,229,255}  
\definecolor{definitioncolor}{RGB}{204,235,204}   
\definecolor{rubriccolor}{RGB}{255,229,255}       
\definecolor{essaycolor}{RGB}{255,255,255}        
\definecolor{outputcolor}{RGB}{250,220,220}       

\begin{document}

\title{Beyond Holistic Scores: Automatic Trait-Based Quality Scoring of Argumentative Essays}

\author{Lucile Favero \inst{1} \orcidID{0009-0005-2981-0124} \and \\ 
Juan Antonio Pérez-Ortiz  \inst{2} \orcidID{0000-0001-7659-8908} \and \\
Tanja Käser \inst{3} \orcidID{0000-0003-0672-0415} \and \\
Nuria Oliver \inst{1} \orcidID{0000-0001-5985-691X}}
\institute{ ELLIS Alicante, Muelle de Poniente 5, Puerto de Alicante, Alicante, 03001, Spain \and Universitat d'Alacant, Carretera San Vicente del Raspeig s/n, San Vicente del Raspeig, Alicante 03690 Spain \and École Polytechnique Fédérale de Lausanne (EPFL), Station 14 Lausanne 1015 Switzerland\\
\email{lucile@ellisalicante.org}}
\titlerunning{Trait-based Automatic Scoring for Argumentative Essays}
\authorrunning{Favero et al.}
\maketitle       
\begin{abstract}
Automated Essay Scoring (AES) systems have traditionally focused on holistic scores, limiting their pedagogical usefulness, especially in the case of complex essay genres such as argumentative writing. In educational contexts, teachers and learners require interpretable, trait-level feedback that aligns with instructional goals and established rubrics. In this paper, we study trait-based Automatic Argumentative Essay Scoring (AAES) using two complementary modeling paradigms designed for realistic educational deployment: (1) structured in-context learning with small open-source large language models (LLMs), and (2) a supervised, encoder-based BigBird model with a CORAL-style ordinal regression formulation, optimized for long-sequence understanding.
We conduct a systematic evaluation with 1,783 argumentative essays from the ASAP++ dataset, which includes essay scores across five quality traits, offering strong coverage of core argumentation dimensions. LLMs are prompted with designed, rubric-aligned in-context examples, along with feedback and confidence requests, while we explicitly model ordinality in scores with the BigBird model via the rank-consistent CORAL framework. 
Our results show that explicitly modeling score ordinality substantially improves agreement with human raters across all traits, outperforming LLMs and nominal classification and regression-based baselines. This finding reinforces the importance of aligning model objectives with rubric semantics for educational assessment. At the same time, small open-source LLMs achieve a competitive performance without task-specific fine-tuning, particularly for reasoning-oriented traits, while enabling transparent, privacy-preserving, and locally deployable assessment scenarios. 
This work advances AAES beyond holistic scoring. Our findings provide methodological, modeling, and practical insights for the design of AI-based educational systems that aim to deliver interpretable, rubric-aligned feedback for argumentative writing.

\end{abstract}

\section{Introduction}
Argumentative writing is a foundational academic and civic skill. It requires learners to formulate defensible claims, support them with relevant evidence, and articulate coherent lines of reasoning. These abilities are closely linked to critical thinking and make argumentative writing assessment a central concern in education \cite{wachsmuth2017argumentation}. Evaluating argumentative quality is demanding, as it requires judgments about reasoning coherence rather than surface linguistic features, making manual assessment time-consuming and variable \cite{wang-etal-2024-beyond-agreement}.

Automated Essay Scoring (AES) has traditionally emphasized holistic scores, limiting its pedagogical usefulness for learning \cite{shermis2013handbook}. In the case of complex genres, such as argumentative writing, holistic assessment hides the multidimensional nature of writing quality and provides little actionable feedback for learners. In contrast, trait-based assessment, evaluating dimensions such as idea development, organization, and language control, is central to formative feedback, instructional alignment, and the development of argumentative competence \cite{wang-etal-2024-beyond-agreement}.

From an Artificial Intelligence in Education (AIED) perspective, this limitation is particularly problematic. Educational applications require assessment models that are interpretable, rubric-aligned, and deployable in real classrooms, supporting teachers and students rather than reproducing summary scores. Despite its importance, trait-based Automatic Argumentative Essay Scoring (AAES) remains underexplored, with most prior work focusing on holistic scoring or on a limited subset of traits.
Recent advances in Large Language Models (LLMs) have opened new possibilities for AES through zero-shot and in-context learning approaches \cite{favero2026argumentative}. However, existing LLM-based studies largely focus on holistic scoring, rely heavily on proprietary systems, raising concerns about transparency, reproducibility, and educational deployment \cite{oketch2025bridging}. In parallel, supervised encoder-based models have demonstrated strong performance for essay scoring, yet typically ignore the ordinal structure of rubric-based scores and remain focused on holistic outcomes \cite{shermis2013handbook}.

While some studies report trait-level AAES results with LLMs \cite{jordan2025magic}, many rely on averaging scores across prompts, genres, rubric definitions, and essay types~\cite{eltanbouly2025trates}. In such settings, identically named traits (\emph{e.g.}, \textit{Content} or \textit{Organization}) may correspond to different evaluative constructs, depending on the prompt provided to write the essay, undermining both interpretability and instructional validity. From an AIED perspective, this practice obscures what is actually being assessed and weakens the pedagogical meaning of trait-level feedback.

In this paper, we address these gaps by studying trait-based AAES using two complementary, educationally realistic modeling paradigms under classroom constraints. We first investigate structured in-context learning with small open-source LLMs, guided by rubric-aligned prompts that elicit justifications and confidence estimates. We then propose a BigBird-based~\cite{zaheer2020big} supervised scoring model using a CORAL-style ordinal regression formulation~\cite{cao2020rank}, explicitly capturing the ordered nature of rubric-based trait scores and aligning model behavior with human grading practices.
We conduct a systematic evaluation on the argumentative subset of the ASAP\texttt{++} dataset,\footnote{\url{https://lwsam.github.io/ASAP++/lrec2018.html}} which provides analytic scores across five core writing traits and strong coverage of argumentative quality dimensions.

By comparing explanation-capable LLM-based scoring with ordinal neural models and standard non-ordinal baselines, we provide empirical and methodological insights into the trade-offs between accuracy, interpretability, and deployability in educational assessment systems.

\begin{figure}[h!]
    \centering
    \includegraphics[width=0.99\linewidth]{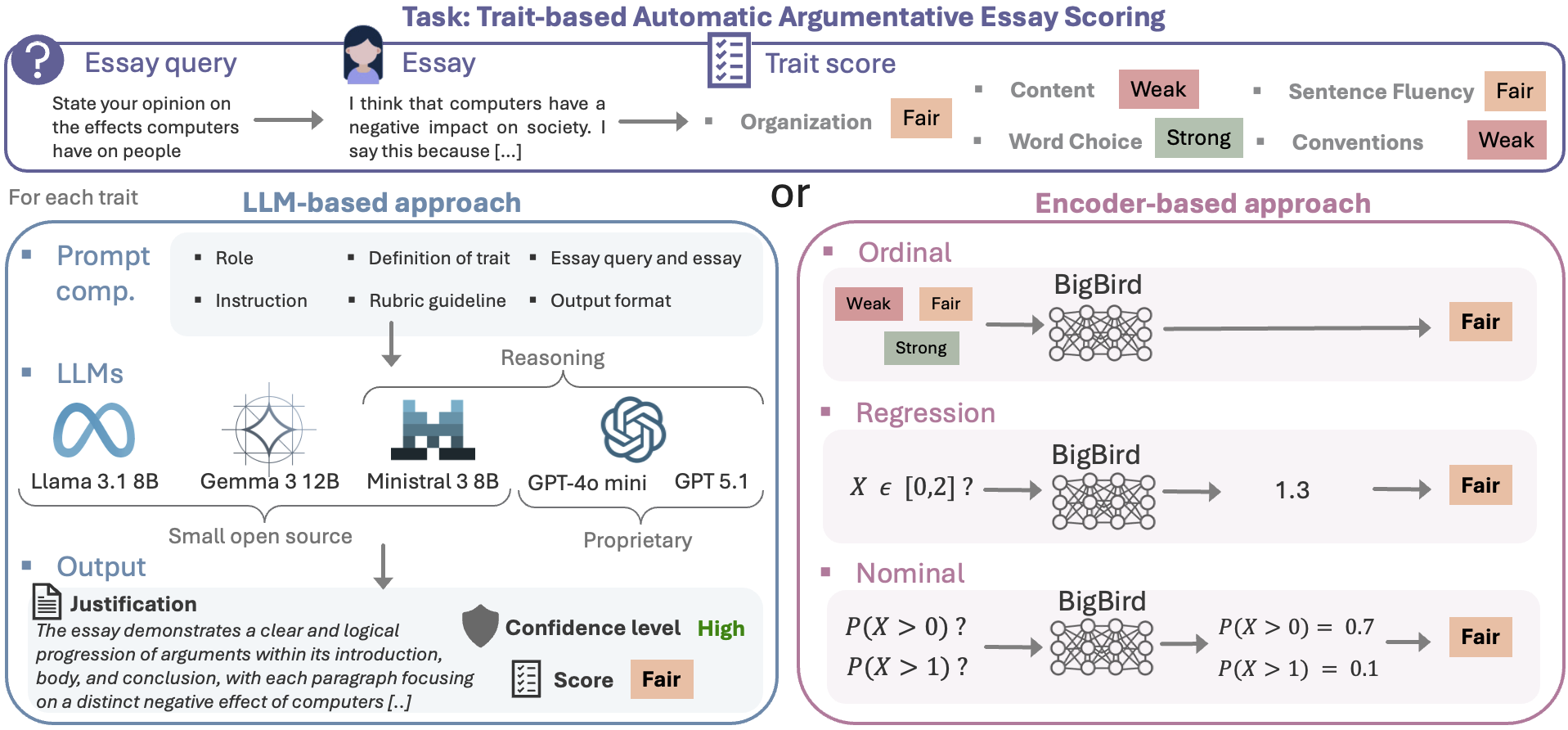}
    \caption{Overview of the two trait-based AAES approaches studied in this paper. }
    \label{fig:archi}
\end{figure}

\paragraph{Contributions} This work makes three key contributions to AIED research: (1) a comparison between small open-source LLMs with structured in-context learning, and supervised, encoder-based models optimized for long-sequence understanding to perform the task of trait-based essay scoring under realistic educational constraints; (2) evidence that explicit ordinal modeling in encoder-based models (CORAL-style) improves agreement with human raters by better aligning with rubric semantics; and (3) empirical evidence supporting the use of small open-source LLMs to generate competitive, interpretable scores, enabling transparent and privacy-preserving deployment in the classroom. Figure~\ref{fig:archi} provides an illustration of the two trait-based AAES approaches studied in this paper. 

\section{Related work}
\paragraph{Automatic Argumentative Essay Scoring (AAES).}
Automatic Essay Scoring (AES) has a long history in educational technology, with early systems relying on surface-level linguistic features to predict holistic scores \cite{shermis2013handbook}. Although effective for large-scale summative assessment, such approaches have been widely criticized for limited construct validity and poor alignment with instructional goals, especially for cognitively demanding genres such as argumentative writing, where quality cannot be reduced to fluency or lexical sophistication alone \cite{wang-etal-2024-beyond-agreement}.
In the specific case of argumentative writing, trait-level scoring  (\emph{e.g.}, claim clarity, evidence relevance, reasoning quality) is particularly important, as it enables the assessment of core argumentation skills and supports formative feedback and more fine-grained learning analytics that can guide the learners' argumentative development \cite{wachsmuth2017argumentation}. 
Although trait-level scoring is increasingly reported \cite{favero2026argumentative}, prior work often aggregates results across essay types to increase sample size or stability \cite{eltanbouly2025trates}. However, because trait definitions are prompt-specific in datasets such as ASAP\texttt{++}, this aggregation implicitly assumes construct equivalence, which does not hold in practice.
Moreover, recent analyses of AAES benchmarks \cite{favero2026argumentative} indicate that few publicly available datasets meaningfully cover core argument quality dimensions. Existing resources often focus on a limited subset of traits, while neglecting other essential aspects of argumentative quality \cite{favero2026argumentative}. This fragmentation constrains the development and evaluation of systems that aim to assess argumentative competence in a pedagogically meaningful way.  

\paragraph{Supervised neural approaches for AES.} 

Supervised neural approaches have substantially advanced AES, particularly through the adoption of transformer-based language models that provide rich contextual representations of student writing \cite{elmassry2025systematic}. BERT-based encoder-only models, for instance, have been fine-tuned to predict essay scores with strong performance on widely used benchmarks, such as ASAP\footnote{\url{https://www.kaggle.com/c/asap-aes} \url{https://lwsam.github.io/ASAP++/lrec2018.html}} and ELLIPSE\footnote{\url{https://github.com/scrosseye/ELLIPSE-Corpus/tree/main}}. Subsequent architectures such as RoBERTa~\cite{liu2019roberta} and ModernBERT~\cite{modernbert}, which improve pretraining objectives and optimization strategies, have further enhanced generalization to complex and linguistically diverse essays.

A critical limitation of early transformer-based AES models, however, is their restricted input length, which limits their ability to model long-form student essays end-to-end \cite{elmassry2025systematic}. This is particularly problematic for argumentative writing, where discourse structure and cross-paragraph dependencies are central to quality judgments \cite{wachsmuth2017argumentation}. To address this challenge, long-document transformers such as Longformer \cite{beltagy2020longformer} and BigBird \cite{zaheer2020big} have been proposed. These models employ sparse attention mechanisms that scale efficiently to longer sequences. In particular, BigBird relies on block-sparse attention and can process sequences of up to 4096 tokens at substantially lower computational cost than standard full-attention transformers, making it well-suited for essay-level modeling.

While many supervised AES systems formulate scoring as either regression or nominal classification \cite{elmassry2025systematic}, such approaches fail to explicitly encode the ordered structure of performance levels. Ordinal regression formulations, such as COnsistent RAnk Logits (CORAL) \cite{cao2020rank}, address this limitation by modeling score thresholds rather than independent class labels. This approach reduces penalization for near-miss predictions and yields more consistent and educationally meaningful score assignments.
 
Transformer-based ordinal models have demonstrated empirical performance and improved alignment with human grading practices. However, most prior work applying ordinal formulations remains focused on holistic essay scoring, with limited attention to analytic, trait-level assessment and to cognitively demanding genres such as argumentative writing \cite{nam2025ordinal,chakravarty-etal-2025-enhancing}. As a result, the potential of ordinal neural modeling for trait-based AAES remains underexplored, particularly in settings that require interpretable and pedagogically grounded assessment.

\paragraph{LLM-based approaches for AAES.} \label{sec:icl_rel_work}
LLMs have recently been explored for automatic argumentative essay scoring using zero-shot and in-context learning (ICL) paradigms, in which task-relevant contextual information is embedded directly in the prompt \cite{jordan2025magic,eltanbouly2025trates}. Prior studies show that carefully designed prompts, incorporating (1) role specification, (2) explicit definitions of argumentative traits, (3) rubric-based scoring criteria, (4) exemplar essays, and (5) structured instructions encouraging step-by-step reasoning, can yield promising agreement with human scores without task-specific fine-tuning \cite{favero2026argumentative}. Recent work has further explored LLMs with reasoning capabilities \cite{oketch2025bridging}, which have been explicitly trained to generate intermediate reasoning steps to improve the reliability, consistency, and interpretability of model outputs.
Nevertheless, existing studies are largely dominated by proprietary models and focus primarily on holistic scoring \cite{favero2025leveraging}, and there is limited prior research on trait-based scoring with LLMs, especially for argumentative writing \cite{jordan2025magic}. Concerns regarding transparency, data privacy, reproducibility, cost, and deployability further limit the applicability of large proprietary models in educational contexts \cite{favero2026argumentative}, 
where interpretability, rubric fidelity, institutional control, and student privacy are essential.

In contrast, our work focuses on trait-based automatic argumentative essay scoring, and compares small open-source LLMs with supervised, encoder-based models,
addressing key gaps in prior work concerning educational validity and practical deployment of AAES methods in learning environments.

\section{Method}

Given an argumentative essay, the aim is to automatically score it according to different argumentative traits.
\subsection{Dataset}
We conduct our experiments on the ASAP\texttt{++} dataset\footnote{\url{https://www.kaggle.com/c/asap-aes}, \url{https://lwsam.github.io/ASAP++/lrec2018.html}}, a widely used benchmark dataset for educational AES research that supports trait-based scoring. The dataset consists of $12,978$ essays of various types (source-dependent, narrative, argumentative) distributed across 8 prompts (or assignment queries).
\texttt{Prompt 1} consists of $1,783$ argumentative essays written by Grade~8 students in response to an argumentative writing prompt asking them to argue their position on the effects of computers on people. The essays are scored on a six-point scale (1--6) across five analytically defined traits: \textit{Ideas and Content:} how fully an essay develops its ideas, how clearly those ideas are expressed, and how effectively they are supported; \textit{Organization:} how well structured the essay is; \textit{Word Choice:} how precisely, effectively, and appropriately a writer selects and uses vocabulary to convey meaning, engage the reader, and suit the audience and purpose of the essay; \textit{Sentence Fluency:} how smoothly and naturally sentences flow in a text, reflecting variety, rhythm, and structural control that make the writing easy and pleasant to read aloud and to follow; and \textit{Conventions:} the writer’s control of standard written English, including punctuation, spelling, capitalization, grammar, and usage, and how effectively these elements support clarity and readability. Essays average approximately $365$ tokens, with some extending up to $785$, making the dataset appropriate for evaluating long-document modeling approaches. 

ASAP\texttt{++} provides a comprehensive coverage of argumentative quality traits when mapped to argumentation theory \cite{favero2026argumentative}. In particular, it is one of the few large-scale, publicly available student essay datasets whose rubric-based traits span multiple Argument Quality (AQ) dimensions---including dialectical reasonableness, deliberative norms, and elements of logical cogency \cite{romberg-etal-2025-towards,favero2026argumentative}---whereas most existing datasets disproportionately emphasize rhetorical and linguistic features  \cite{favero2026argumentative}. This makes ASAP\texttt{++} especially suitable for studying argumentative reasoning rather than surface writing quality. See the \href{https://anonymous.4open.science/r/AIED2026-E0DB/AIED_2026_AAES_annex.pdf}{online Annex B} for a comprehensive mapping of each trait to the AQ theory.  

In contrast to common practice in trait-level AES, we do not average the traits across essay types. In ASAP\texttt{++}, analytic traits are defined relative to prompt-specific rubrics, and their operationalization varies across genres. Averaging trait scores across prompts would therefore conflate distinct constructs under a shared label, weakening both interpretability and educational validity. Because our goal is to study argumentation-specific trait assessment, we restrict our analysis to the argumentative prompt (\texttt{Prompt 1}) and treat trait scores as prompt-bound constructs.

To improve both educational relevance and modeling robustness, we turn the original six-point score scale into three ordinal categories by mapping scores 1--2, 3--4, and 5--6 into low, medium, and high performance levels, respectively. Fine-grained six-point distinctions are often of limited pedagogical value for students and may introduce artificial precision in formative assessment \cite{linacre2015sin}.
From a modeling perspective, LLMs are less reliable when required to discriminate between narrowly defined ordinal categories without clear semantic boundaries. We therefore assign semantically meaningful labels---\textit{weak}, \textit{fair}, and \textit{strong}---instead of a purely numeric scale. Such semantically grounded categories have been shown to improve model performance and alignment with human judgments, while enhancing interpretability in educational assessment settings \cite{favero2026argumentative}.
    
\subsection{LLMs for AAES} 

Recent work in automatic argumentative essay scoring with large language models is largely dominated by proprietary systems \cite{favero2026argumentative}. In contrast, we emphasize small\footnote{We denote by small LLMs those with less than 13B parameters because they are deployable on commodity, off-the-shelf hardware.} open-source LLMs, which are particularly relevant to educational settings due to several key properties: (1) they enable local or institutional deployment, supporting data privacy and regulatory compliance, transparency and inspection; (2) they allow for task-specific adaptation aligned with pedagogical goals; and (3) they can be deployed in off-the-shelf computers at affordable cost \cite{oketch2025bridging}. Moreover, smaller models are less prone to over-authoritative behavior and are easier to tune for educationally appropriate feedback and assessment.
AES places higher demands on reasoning than non-argumentative AES tasks, as it requires modeling argumentative structure and justification rather than surface-level writing quality alone \cite{wachsmuth2017argumentation}. To examine the role of reasoning in this context, we compare two well-established non-reasoning small open-source models (Llama-3.1 8B \cite{grattafiori2024llama}, and Gemma-3 12B \cite{team2025gemma}), with a reasoning-oriented small open-source model (Ministral-3 8B \cite{liu2026ministral}).  

Moreover, we contextualize the performance of open-source models relative to current proprietary approaches by including a small proprietary reasoning model, GPT-4o-mini (8B parameters), and a state-of-the-art one, GPT-5.1 \cite{achiam2023gpt}, which is estimated to have at least two orders of magnitude more parameters.

\paragraph{Model configuration and inference settings.}
In our experiments, the LLMs we deployed by hosting them locally in \textit{Ollama}\footnote{\url{https://github.com/ollama}, \url{https://ollama.com}} to ensure full control over inference conditions and data locality. The models were configured with the default temperature value ($0.8$) and a maximum generation length of $10,000$ tokens, allowing them to produce extended, coherent outputs suitable for reasoning in argumentative essay scoring. To avoid residual state effects across interactions, we turned off session persistence and caching, ensuring that each model invocation was independent and reproducible. 
\paragraph{In-Context Learning (ICL).}

In line with the recent findings on ICL in AAES (see Section~\ref{sec:icl_rel_work}), we designed a structured prompt (see \href{https://anonymous.4open.science/r/AIED2026-E0DB/AIED_2026_AAES_annex.pdf}{online Annex C}) to guide the LLMs toward reliable, trait-consistent scoring. The prompt is composed of:

\begin{itemize}
    \item \colorbox{orange!30}{Role specification}: The model is instructed to act as an expert evaluator of the students' argumentative essays, following established scoring rubrics. It is explicitly asked to be fair, objective, and consistent across evaluations.
    
    \item \colorbox{instructioncolor}{Task instruction}: The prompt provides contextual information about the essay genre (argumentative writing) and the target population (13-year-old students). This information is intended to calibrate the evaluation criteria to age-appropriate expectations. The scoring task is explicitly described and presented in accessible terms to reduce ambiguity.
    
    \item \colorbox{definitioncolor}{Trait definition}: A dedicated section defines the evaluated trait, aligned with the official guidelines of the dataset, ensuring conceptual consistency between human and automated scoring.
    
    \item \colorbox{rubriccolor}{Rubric guidelines with exemplars}: For each score level, the prompt includes a concise description of the corresponding essay characteristics for the given trait, summarised with the official guidelines of the dataset (see \href{https://anonymous.4open.science/r/AIED2026-E0DB/AIED_2026_AAES_annex.pdf}{online Annex A}). Each level is accompanied by an essay as an example to provide concrete reference points and support ICL. To mitigate potential data leakage, all in-context exemplars were not test essays.
    
    \item \colorbox{lightgray!40}{Essay query}: The student's essay to be evaluated is presented clearly and separated from the instructional content to avoid confusion between examples and the target input.
    
    \item \colorbox{outputcolor}{Output specification}: The model is instructed to (i) justify the assigned score, (ii) provide the final score label, and (iii) report a confidence level for the decision. The response must be in JSON format to ensure clear separation between reasoning and the final output, facilitating automatic parsing and evaluation. 
\end{itemize}

An example of an LLM output can be found in the \href{https://anonymous.4open.science/r/AIED2026-E0DB/AIED_2026_AAES_annex.pdf}{online Annex D}.
Note that small LLMs have been found to be sensitive to prompt wording and to the relative positioning of each component within the overall prompt \cite{zhuo-etal-2024-prosa,guan2025order}. For instance, when the \textit{Output specification} section is not placed at the very end of the prompt, the model frequently fails to comply with the required JSON output format. 
Prompt length is another critical factor, especially given the relatively long input essays considered in this work. Increasing the number of in-context essay exemplars beyond a single example per score level often leads to malformed outputs and a substantial degradation in scoring performance. These observations highlight the importance of careful prompt engineering when applying ICL with small LLMs in educational assessment settings \cite{jordan2025magic}.

\subsection{BigBird CORAL for AAES}

 BigBird is an encoder-only transformer-based model with block-sparse attention to handle long sequences efficiently, addressing one of the key limitations of the original models like BERT \cite{zaheer2020big}. We adopt it as our backbone model, as it has demonstrated superior performance in AAES tasks when compared to other encoder-only approaches \cite{nam2025ordinal}. The contextualized representation of the first token is extracted from the final hidden layer, passed through a dropout layer, and projected via a linear layer to produce one logit per ordinal threshold. The model is fine-tuned end-to-end for each trait dimension.

\paragraph{CORAL: Ordinal formulation.}
Rather than modeling the task as a nominal 3-way classification problem, we formulate automated essay scoring as an \emph{ordinal regression} task using a CORAL-style thresholding approach \cite{cao2020rank}. For the three ordered score categories---weak (1), fair (2), and strong (3)---the model predicts two threshold outcomes corresponding to the probability that the true label $y$ exceeds each boundary, namely $P(y>1)$ and $P(y>2)$. This formulation explicitly encodes the ordered nature of rubric-based scores and reduces the penalization of near-miss predictions compared to nominal classification \cite{nam2025ordinal,cao2020rank}.

Because ordinal threshold models require mapping threshold probabilities to discrete score labels, we additionally tune the decoding cutoffs to optimize QWK, a standard agreement metric in educational assessment (see Section~\ref{sec:mth_metric}). Specifically, for our two-threshold setting, we perform a grid search over cutoff pairs $(c_1, c_2)$ on the validation set only, enforcing monotonicity $(c_2 \ge c_1)$. 

\paragraph{Training objective and class imbalance.}
For each essay with ordinal label $y \in \{1,2,3\}$, we construct a binary target vector of length two indicating whether the label exceeds each threshold, i.e., $[\mathbb{1}(y>1), \mathbb{1}(y>2)]$. Training minimizes binary cross-entropy with logits over these two targets. To address class imbalance, we compute a per-threshold positive class weight from the training data and include it into the loss function, reducing bias toward the majority class.

\paragraph{Data splits.}
We use train, validation, and test splits ($1069$, $357$, and $357$ essays, respectively). Hyperparameter and model selection are performed without access to test labels. Any procedure that depends on ground-truth labels---such as decision threshold optimization---is conducted exclusively on the validation set, and the resulting parameters are subsequently applied to the test set.

\paragraph{Model selection and early stopping.}
Models are trained for up to six epochs, with evaluation and checkpointing performed at fixed step intervals. The best-performing checkpoint is selected based on validation-tuned QWK (see Section \ref{sec:mth_metric}), computed only on the validation split. To mitigate overfitting, we employ early stopping with a patience of two evaluation steps.

\subsection{Baseline BigBird variants}
To analyze the contribution of ordinal modeling, we compare BigBird–CORAL against two baselines: BigBird with nominal classification and BigBird with regression-based scoring. Both represent standard modeling choices that do not explicitly encode the ordered semantics of rubric-based scores \cite{elmassry2025systematic}.

First, we implement a BigBird-based trait scoring model formulated as a three-way classification task (BigBird-nominal). For each trait, the model predicts a categorical label using a softmax output layer, trained with cross-entropy loss. Score categories are treated as nominal, implicitly assuming equal severity for all misclassifications and ignoring the ordinal structure of educational rubrics.

As a second baseline, we use BigBird with a regression objective (BigBird-RMSE) \cite{nam2025ordinal}. Trait scores are mapped to scalar targets, and the model is trained to minimize mean squared error. Continuous predictions are discretized into ordinal categories using fixed thresholds at inference time. Regression captures coarse score ordering but optimizes numerical proximity rather than rubric-consistent thresholds, potentially obscuring near-miss errors that are pedagogically less consequential.
\subsection{Evaluation metrics} \label{sec:mth_metric}
Model performance is primarily evaluated using quadratic weighted kappa (QWK), the standard metric in AES \cite{elmassry2025systematic}. QWK ranges from 0 to 1 and measures agreement between predicted and reference scores while penalizing larger disagreements more strongly. We adopt the conventional interpretation of agreement levels proposed by \cite{landis1977measurement}: \emph{poor} agreement ($0\leq$ QWK $<0.2$); \emph{fair} agreement ($0.2 \leq$ QWK $<0.4$); \emph{moderate} agreement ($0.4 \leq$ QWK $<0.6$); \emph{substantial} agreement ($0.60 \leq$ QWK $<0.80$); and \emph{almost perfect} agreement (QWK $\geq 0.80$).
While widely used, QWK has known limitations for educational assessment: it is sensitive to score distributions and scale definitions, it may be affected by prevalence effects, and it imposes a fixed quadratic penalty that does not necessarily reflect rubric semantics or pedagogical relevance \cite{doewes2023evaluating}. 


We complement QWK with weighted F1, which addresses performance by averaging the F1 scores of each class, weighted by class support, so it reflects a precision–recall balance while accounting for class imbalance, and Kendall’s $\tau$, a non-parametric rank-based correlation that captures ordinal consistency (see \href{https://anonymous.4open.science/r/AIED2026-E0DB/AIED_2026_AAES_annex.pdf}{online Annex E}) These 3 metrics provide a comprehensive view of system performance by combining agreement, ordinal alignment, and score-level accuracy \cite{favero2026argumentative}. 

\section{Results}

To account for stochasticity during training, each experiment is repeated across multiple random seeds, and results are reported per dimension, along with aggregate statistics. The code to run the experiments can be found \href{https://anonymous.4open.science/r/AIED2026-E0DB/README.md}{here}. See also \href{https://anonymous.4open.science/r/AIED2026-E0DB/AIED_2026_AAES_annex.pdf}{online Annex E} for complementary results.


\begin{table}[h!]
\centering
\small
\caption{Performance on \texttt{Prompt 1} of the ASAP\texttt{++} dataset for five traits. Results are averaged over six repetitions (sampling for LLMs and fine-tuning with different seeds for the BigBird models) on the test set. Bold values indicate the best performance overall, while underlined values indicate the best performance among LLMs.}
\begin{tabular}{l|ccccc}
\hline
\textbf{Method} &
\textbf{Content} &
\makecell{\textbf{Organization}} &
\makecell{\textbf{Word}\\\textbf{choice}} &
\makecell{\textbf{Sentence}\\\textbf{fluency}} &
\makecell{\textbf{Conventions}} \\
\hline
\multicolumn{6}{c}{\textbf{QWK}} \\
\hline
Llama-3.1 8B & 0.22 ± 0.03 & 0.25 ± 0.04 & 0.26 ± 0.03 & \underline{0.20 ± 0.04} & 0.22 ± 0.04 \\
Gemma-3 12B & 0.09 ± 0.05& 0.16 ± 0.07 & 0.13 ± 0.07 & 0.13 ± 0.06 & 0.19 ± 0.06 \\
Ministral-3 8B & \underline{0.49 ± 0.07} & \underline{0.43 ± 0.04} & \underline{0.32 ± 0.07} & 0.19 ± 0.04 & \underline{0.33 ± 0.03} \\
BigBird-nominal & 0.39 ± 0.05 & 0.44 ± 0.04 & 0.43 ± 0.04 & 0.33 ± 0.02 & 0.33 ± 0.03 \\
BigBird-RMSE & 0.53 ± 0.02 & 0.49 ± 0.02 & 0.56 ± 0.02 & \textbf{0.47 ± 0.02}& 0.45 ± 0.01\\
BigBird-CORAL & \textbf{0.59 ± 0.01} & \textbf{0.53 ± 0.02} & \textbf{0.61 ± 0.02} & \textbf{0.47 ± 0.01}& \textbf{0.48 ± 0.01} \\
\hline
\multicolumn{6}{c}{\textbf{Weighted F1}} \\
\hline
Llama-3.1 8B & 0.49 ± 0.02 & 0.53 ± 0.01 & 0.61 ± 0.02 & 0.52 ± 0.01 & 0.60 ± 0.02 \\
Gemma-3 12B & 0.59 ± 0.07 & \underline{0.65 ± 0.07} & 0.64 ± 0.03 & 0.62 ± 0.01 & 0.63 ± 0.02 \\
Ministral-3 8B & \underline{0.69 ± 0.04} & 0.63 ± 0.03 & \underline{0.70 ± 0.02} & \underline{0.67 ± 0.03} & \underline{0.70 ± 0.03} \\
BigBird-nominal & \textbf{0.73 ± 0.02} & \textbf{0.74 ± 0.02}& \textbf{0.77 ± 0.02} & \textbf{0.71 ± 0.01} & \textbf{0.71 ± 0.01}\\
BigBird-RMSE & \textbf{0.73 ± 0.01}& \textbf{0.74 ± 0.01} & \textbf{0.77 ± 0.02}& \textbf{0.71 ± 0.01}& \textbf{0.71 ± 0.02} \\
BigBird-CORAL & 0.71 ± 0.03 & 0.68 ± 0.02 & 0.76 ± 0.02 & 0.65 ± 0.01 & 0.67 ± 0.01 \\
\hline
\end{tabular}
\label{tab:results_overview}
\end{table}

Table~\ref{tab:results_overview} compares the small open-source models (reasoning and non-reasoning) against the proposed BigBird-CORAL ordinal model and the two baselines. 

Across all five traits, BigBird with the ordinal formulation (CORAL) consistently achieves the strongest agreement with human raters, outperforming all the LLMs and other variants of BigBird. This pattern confirms the importance of explicitly modeling score ordinality in rubric-based assessment. The non-ordinal variants of BigBird achieve strong weighted F1 scores, yet from an educational perspective, QWK, which measures alignment with human judgments, is a more meaningful metric than F1.

Among the LLMs, Ministral-3~8B consistently outperforms the other models across traits. 
LLM performance varies systematically by trait: the strongest agreement corresponds to \textit{Content} and \textit{Organization}, which emphasize global argumentative reasoning, idea development, and discourse structure. In contrast, agreement is low for \textit{Word Choice}, \textit{Sentence Fluency} 
and \textit{Conventions}, traits 
that depend heavily on surface-level linguistic control and fine stylistic distinctions. This pattern mirrors known challenges in human scoring reliability for language mechanics, and supports the interpretation that model–human disagreement in these traits reflects intrinsic ambiguity rather than model failure \cite{favero2026argumentative}.

The confidence values reported by the models were predominantly high (in the range of $0.8$--$1.0$).
While these values should not be interpreted as calibrated probabilities, they provide a coarse indicator of model certainty that may be useful in educational settings, for instance, to flag low-confidence cases for human review. Importantly, we observed that explicitly requesting both feedback and confidence estimates in the prompt consistently improved scoring performance compared to ablations without these components. This suggests that requiring models to articulate justifications and reflect on their certainty encourages more structured reasoning and more stable, rubric-consistent predictions.

\begin{table}[h!]
\centering
\caption{Trait-based performance on the ASAP\texttt{++} dataset with five traits. Comparison of the best-performing small open-source LLM (Ministral-3 8B) with GPT-4o-mini and GPT-5.1. Results are averaged across three independent runs on the test set. Best results are shown in bold.}
\small
\renewcommand{\arraystretch}{1}
\begin{tabular}{l|l|c|c|c}
\hline
\textbf{Trait} & \textbf{Metric} & \textbf{Ministral-3 8B} & \textbf{GPT-4o-mini} & \textbf{GPT-5.1} \\\hline
Content
& QWK & \textbf{0.46 ± 0.09} & 0.29 ± 0.00 & 0.45 ± 0.02 \\
& Weighted F1 & \textbf{0.69 ± 0.04} & 0.52 ± 0.01 & 0.68 ± 0.01 \\\hline
Organization
& QWK & \textbf{0.42 ± 0.05} & 0.28 ± 0.02 & 0.38 ± 0.02 \\
& Weighted F1 & \textbf{0.64 ± 0.03} & 0.62 ± 0.01 & 0.52 ± 0.01 \\\hline
Word choice
& QWK & 0.29 ± 0.09 & 0.23 ± 0.02 & \textbf{0.42 ± 0.02} \\
& Weighted F1 & \textbf{0.70 ± 0.01} & 0.62 ± 0.02 & 0.66 ± 0.02 \\\hline
Sentence fluency
& QWK & 0.19 ± 0.05 & 0.26 ± 0.02 & \textbf{0.34 ± 0.02} \\
& Weighted F1 & \textbf{0.68 ± 0.05} & 0.55 ± 0.01 & 0.60 ± 0.02 \\\hline
Conventions
& QWK & \textbf{0.33 ± 0.02} & 0.22 ± 0.04 & 0.29 ± 0.01 \\
& Weighted F1 & \textbf{0.71 ± 0.04} & 0.55 ± 0.02 & 0.54 ± 0.02 \\\hline
\end{tabular}
\label{tab:gpt}
\end{table}

Table~\ref{tab:gpt} reports trait-based performance for the best-performing small open-source LLM (Ministral-3~8B) compared with two proprietary models (GPT-4o-mini and GPT-5.1). 
Overall, results indicate that a small open-source LLMs can achieve performance comparable to, and in several cases better than, that of proprietary models across multiple argumentative writing traits, despite the absence of task-specific fine-tuning. GPT-5.1 only outperforms Ministral-3~8B in the QWK of two traits (\textit{Word Choice} and \textit{Sentence Fluency}), suggesting sensitivity to subtle stylistic and rhythmic differences. 
This pattern indicates that argumentative traits that rely more on surface-level linguistic variation may benefit from greater model capacity, while smaller models remain effective for stable, rubric-consistent scoring.

For the QWK metric, paired t-tests revealed significant ($p<0.05$) performance differences across models. BigBird-CORAL significantly outperformed all LLMs and BigBird-nominal on all traits, and significantly outperformed BigBird-RMSE on all traits but \textit{Organization} and \textit{Sentence Fluency}. Ministral-3 8B significantly outperformed Llama-3.1 and Gemma-3 on all traits except \textit{Word Choice} and \textit{Sentence Fluency}. In contrast, there were no significant differences in the performance between Ministral-3 8B and GPT-5.1, except for \textit{Sentence Fluency}, highlighting the competitive performance of Ministral-3 8B despite its much smaller size ($\sim$two orders of magnitude smaller than GPT-5.1 \footnote{\url{https://www.cometapi.com/how-many-parameters-does-gpt-5-have}}). 

\section{Discussion}
From our experiments, we draw several implications for the design of trait-based AAES systems. 

First, we find that BigBird-CORAL  outperforms both all LLMs and the nominal classification and regression baselines on ordinal agreement metrics, 
in contrast to prior work on holistic scoring with BigBird-based ordinal models, which reported no clear gains over regression objectives \cite{nam2025ordinal}. Our results 
illustrate the effectiveness of ordinal modeling at the trait level, where rubric thresholds correspond to interpretable performance bands. For educational use, ordinal consistency is more closely aligned with pedagogical validity than strict category accuracy, supporting the adoption of ordinal formulations for trait-based AAES.

Second, higher agreement with humans is observed in reasoning-oriented traits (\textit{Content}, \textit{Organization}) than stylistic traits (\textit{Sentence Fluency}, \textit{Conventions}), a pattern consistent with known variability in human scoring \cite{shermis2013handbook}. This finding suggests that lower agreement on certain traits reflects intrinsic construct ambiguity rather than model failure. 

Third, the best-performing small open-source LLM (Ministral-3 8B) achieves competitive performance, even when compared to GPT-5.1, a significantly larger proprietary model, particularly 
on \textit{Content}, \textit{Organization} and \textit{Conventions}. This finding suggests that structured, rubric-aligned prompting enables small models to reason effectively about higher-level argumentative quality. From an education perspective, this finding supports their use for privacy-preserving and locally deployable AIED systems \cite{favero2025leveraging}.

\section{Conclusion}
In this paper, we have focused on trait-based argumentative essay scoring and compared small, open-source LLMs that employ structured in-context learning with supervised BigBird-based models. We show that different modeling paradigms offer complementary strengths for educational assessment.
BigBird-CORAL achieves the highest agreement with human raters, whereas small open-source LLMs demonstrate competitive performance, even outperforming proprietary LLMs.

Our results highlight that trait-based AAES, supported by ordinal modeling and transparent LLM prompting, represents a promising step toward scalable, learning-centered support for argumentative writing.

Despite the competitive results, our work is not exempt from limitations. Notably, our study is limited to a single argumentative prompt and focuses on scoring accuracy rather than downstream learning effects. Future work should examine cross-prompt generalization and empirically evaluate how students and teachers could use and benefit from trait-level scores in real-world classroom settings. Moreover, we plan to investigate hybrid assessment frameworks that integrate the stability and higher agreement of ordinal models with the explanations, confidence estimates, and revision-oriented feedback of LLMs.

\section*{Acknowledgements}
This work has been partially supported by a nominal grant received at the ELLIS Unit Alicante Foundation from the Regional Government of Valencia in Spain (Resolución of the Generalitat Valenciana, Conselleria de Innovación, Industria, Comercio y Turismo, Dirección General de Innovación). L.F. has also been partially funded by the Bank Sabadell Foundation.


\end{document}